\documentclass[letterpaper, 10 pt, conference]{ieeeconf}  

\IEEEoverridecommandlockouts                              
\usepackage{fourier} 
\usepackage{array}
\usepackage{makecell}
\usepackage{dblfloatfix}

\usepackage{multibib}
\usepackage{multirow}
\usepackage{pifont}

\usepackage{lipsum}
\usepackage{subcaption}
\usepackage{amssymb,amsmath}

\usepackage{graphicx}
\usepackage{url}
\usepackage{cite} 
\usepackage{booktabs}
\usepackage{tabularx}

\usepackage[usenames,dvipsnames]{color}
\definecolor{dark-gray}{gray}{0.30}
\definecolor{orange}{rgb}{1.0,0.5,0}

 \newcommand {\argmax}[1]{\underset{#1}{\operatorname{argmax}}}

\DeclareRobustCommand*{\IEEEauthorrefmark}[1]{%
  \raisebox{0pt}[0pt][0pt]{\textsuperscript{\footnotesize\ensuremath{#1}}}}

\title{\textsc{Mathematical Models of Adaptation \\ in Human-Robot Collaboration}
 \\\vspace{0.3cm}
 \large{Stefanos Nikolaidis\IEEEauthorrefmark{1}, Jodi  Forlizzi\IEEEauthorrefmark{2}, David Hsu\IEEEauthorrefmark{3},  Julie Shah\IEEEauthorrefmark{4} and Siddhartha Srinivasa\IEEEauthorrefmark{1} \\ \vspace{0.3cm}
\small\IEEEauthorrefmark{1}The Robotics Institute, Carnegie Mellon University \\\IEEEauthorrefmark{2}Human-Computer Interaction Institute, Carnegie Mellon University\\
  \IEEEauthorrefmark{3}Department of Computer Science, National University of Singapore\\\vspace{-0.15cm}
    \IEEEauthorrefmark{4}Computer Science and Artificial Intelligence Laboratory, Massachusetts Institute of Technology, \thanks{This is a summary of work done in collaboration with Keren Gu, Anton Kuznetsov, Minae Kwon, Premyszlaw Lasota, Swaprava Nath, Ariel Procaccia, Ramya Ramakrishnan, Yu Xiang Zhu.}
    }\\\vspace{0.1cm}\texttt{\small snikolai@cmu.edu, forlizzi@cs.cmu.edu, dyhsu@comp.nus.edu.sg},\vspace{-0.15cm}\\\texttt{\small julie\_a\_shah@csail.mit.edu, siddh@cmu.edu}\vspace{-0.6cm}
}

\begin{document}

\maketitle

\begin{abstract}
A robot operating in isolation needs to reason over the uncertainty in its model of the world and \emph{adapt} its own actions to account for this uncertainty. Similarly, a robot interacting with people needs to reason over its uncertainty over the human internal state, as well as over how this state may change, as humans \emph{adapt} to the robot. This paper summarizes our own work in this area, which depicts the different ways that probabilistic planning and game-theoretic algorithms can enable such reasoning in robotic systems that collaborate with people. We start with a general formulation of the problem as a two-player game with incomplete information. We then articulate the different assumptions within this general formulation, and we explain how these lead to exciting and diverse robot behaviors in real-time interactions with actual human subjects, in a variety of manufacturing, personal robotics and assistive care settings.
\end{abstract}

\section{Introduction}

In human collaboration, the success of the team often depends on the ability of team members to coordinate their actions, by reasoning over the beliefs and actions of their teammates. We want to enable robot teammates with this very capability in human-robot teams, e.g., service robots interacting with users at home, manufacturing robots sharing the same physical scape with human mechanics and autonomous cars interacting with drivers and pedestrians.

When it comes to robots operating in isolation, there has been tremendous progress in enabling them to act autonomously by reasoning over the physical state of the world. A manipulator picking up a glass needs to know the position and orientation of the glass on the table, the location of other objects that it should avoid, and the way these objects will move if pushed to the side. More importantly, it needs to reason over the uncertainty in its model of the world and \textit{adapt} its own actions to account for this uncertainty, for instance by looking at the table with its camera, or by moving slowly until it senses the glass in its gripper.

\begin{figure}[t!]
\centering
  \includegraphics[width=0.8\linewidth]{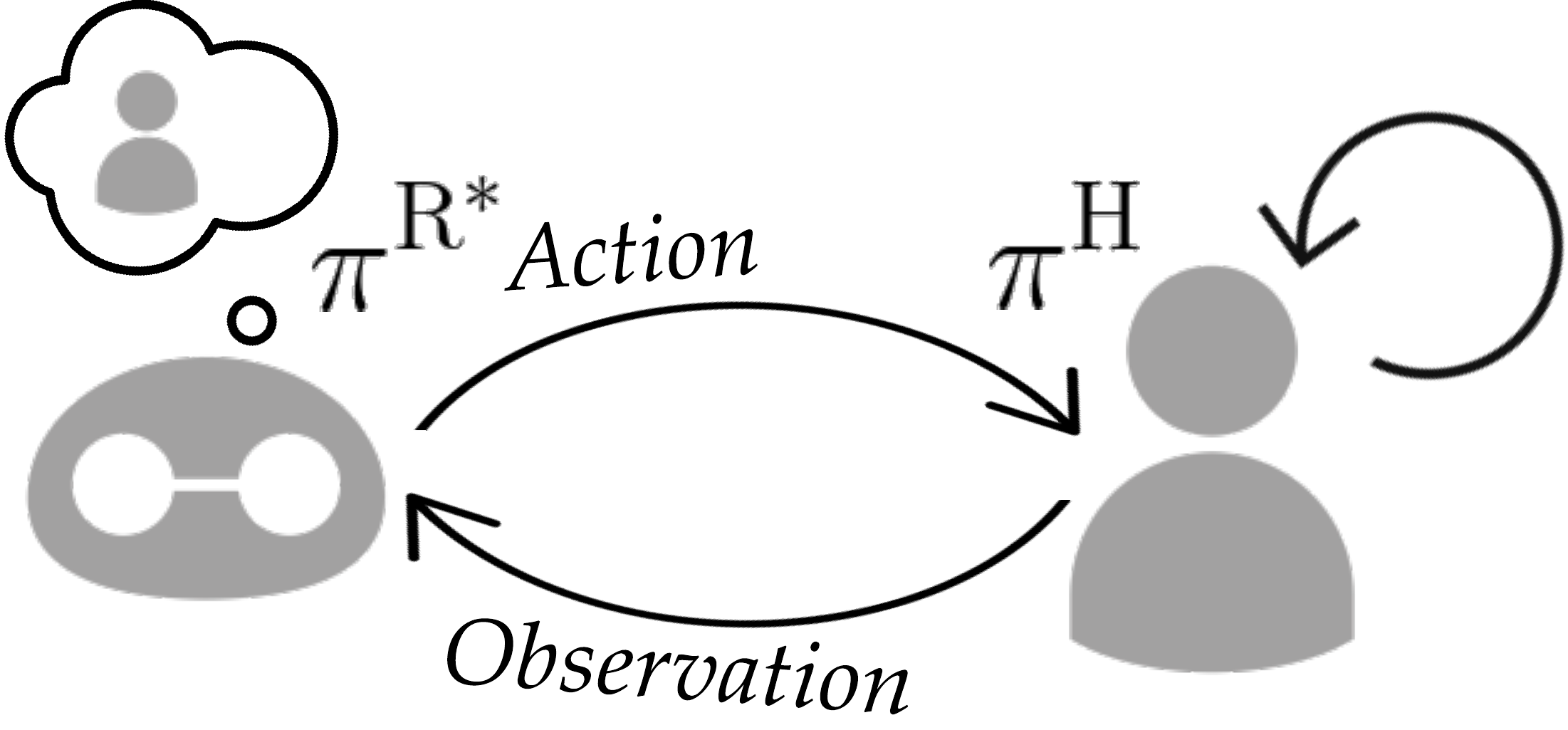}
 \caption{The robot maximizes the performance of the human-robot team by executing the optimal policy $\pi^{\textrm{R}^*}$. The robot takes \emph{information-seeking} actions that allow estimation of the human policy $\pi^{\textrm{H}}$, but also \emph{communicative} actions that guide $\pi^{\textrm{H}}$ towards better ways of doing the task. These actions emerge out of optimizing for $\pi^{\textrm{R}^*}$.}
 \label{fig:diagram}
\end{figure}

 However, humans are not just obstacles that the robot should avoid. They are intelligent agents with their own internal state, i.e., their own goals and expectations about the world and the robot. Their state can change, as they \textit{adapt} themselves to the robot and its actions. Much like in manipulation, a robot interacting with people needs to use this information when choosing its own actions (Fig.~\ref{fig:diagram}).  This requires not only an understanding of human behavior when interacting with robotic systems, but also of the computational challenges and opportunities that arise by enabling this reasoning into deployed systems in the real world.  

 To address these challenges, we have used insights from behavioral economics to propose scalable models of human behavior and machine learning algorithms to automatically learn these models from data. Integrating these models into probabilistic planning and game-theoretic algorithms has allowed generation of robot actions in a computationally tractable manner. 

\begin{figure*}[t!]
\centering
\includegraphics[height=3.2cm]{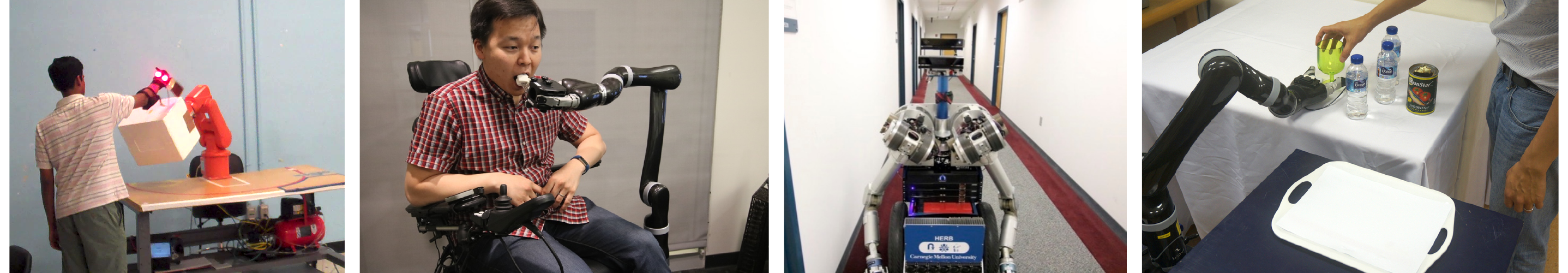} \\
\caption{We have applied our research to human-robot collaboration across different robot morphologies and settings: in manufacturing, assistive care, social navigation and at home.}
\label{fig:applications}
\end{figure*}

This paper is a summary of our work in this area, which has been inspired by recent technical advances in human-robot interaction~\cite{thomaz_IJRR_computational}, and to a large extent it has been made possible by breakthroughs in computational representations of uncertainty~\cite{ong_IJRR_planning}, and in algorithms that have leveraged these representations~\cite{kurniawati_RSS_sarsop}. It has also been inspired by  the remarkable results of game-theoretic algorithms in deployed applications~\cite{pita_2009_losangeles}.

We start by formulating our overarching goal of computing the robot actions that maximize team performance as an optimization problem in a two-player game with incomplete information (Sec.~\ref{sec:formulation}). In hindsight, our approaches in the last years reflect the different assumptions and approximations that we made within the scope of this general formulation.

\renewcommand\theadalign{cb}
\renewcommand\theadfont{\bfseries}
\renewcommand\theadgape{\Gape[4pt]}
\renewcommand\cellgape{\Gape[4pt]}
\newcolumntype{P}[1]{>{\centering\arraybackslash}p{#1}}
\newcolumntype{M}[1]{>{\centering\arraybackslash}m{#1}}

\begin{table*}[b!]%
\centering
\begin{tabular}{ m{2.5cm}  P{2cm}   P{2cm}  P{2cm}  P{2cm}  P{2cm}  P{2cm} }
    
\multicolumn{2}{  c  }{}
 & \thead{Cross\\Training\\\cite{nikolaidis_IJRR_crosstraining, nikolaidis_HRI_crosstraining, nikolaidis_ISR_collaboration}} & \thead{Human \\ Type Inference\\\cite{nikolaidis_HRI_learningtypes} } & \thead{Bounded \\ Memory \\\cite{nikolaidis_HRI_mutualadaptation, nikolaidis_IJRR_mutualadaptation}} & \thead{Best\\Response\\\cite{nikolaidis_HRI_gametheory}} & \thead{Mutual\\Adaptation\\\cite{nikolaidis_HRI_mutualadaptation, nikolaidis_HRI_sharedautonomy, nikolaidis_IJRR_mutualadaptation}}\\   \\
 \multirow{2}{*}{\thead{\vspace{-0.25cm}\normalfont{\textsc{Human Model}}}} & \thead{\normalfont\itshape{Known}} & & & \large\ding{51} & \large\ding{51} & \\ & \thead{\normalfont\itshape{Unknown}}& \large\ding{51} & \large\ding{51} & & & \large\ding{51}\\  \\
 \multirow{2}{*}{\thead{\vspace{-0.25cm}\normalfont{\textsc{Interaction Model}}}} & \thead{\normalfont\itshape{Independent}} &  \large\ding{51} & \large\ding{51} & & & \\ & \thead{\normalfont\itshape{Underactuated}} & & & \large\ding{51} & \large\ding{51} & \large\ding{51} \\ \\
\multirow{2}{*}{\thead{\vspace{-0.25cm}\normalfont{\textsc{Teamwork Model}}}} & \thead{\normalfont\itshape{Leader-Assistant}} &  \large\ding{51} & \large\ding{51} & & &\\ 
 & \thead{\normalfont\itshape{Equal Partners}} & & &  \large\ding{51}& \large\ding{51} & \large\ding{51}\\  \\ 
\end{tabular}
\caption{Each work makes different assumptions on the human, interaction and teamwork models.}
\label{table:summary}
\end{table*}

We start by assuming a \emph{leader-assistant} teamwork model~\footnote{We refer the reader to~\cite{shah2011improved} for the definitions of \emph{leader-assistant} and \emph{equal partners} teamwork models.}, where the goal of the robot is to follow the human preference (Sec.~\ref{sec:robot_adaptation}). We show that this model is an instance of our general framework by representing the human preference as a reward function, shared by both agents and unknown to the robot. We show that cross-training~\cite{nikolaidis_HRI_crosstraining, nikolaidis_IJRR_crosstraining}, an algorithm emulating effective human team training practices, as well as clustering users into types from joint-action demonstrations~\cite{nikolaidis_HRI_learningtypes} significantly improves \emph{robot adaptation} to the human.

We then examine the case when the robot can indirectly affect human actions as an \emph{equal partner}, by treating the interaction as an underactuated dynamical system (Sec.~\ref{sec:human_adaptation}). We present a \emph{bounded-memory model}~\cite{nikolaidis_HRI_mutualadaptation, nikolaidis_HRI_sharedautonomy, nikolaidis_IJRR_mutualadaptation} and a \emph{best-response model} of human behavior~\cite{nikolaidis_HRI_gametheory}, and show that this results in \emph{human-adaptation} to the robot. 

Finally, closing the loop between the two results in \emph{mutual adaptation} (Sec.~\ref{sec:human_adaptation}): The robot builds online a model of the human adaptation by taking information seeking actions, and adapts its own actions in return~\cite{nikolaidis_HRI_mutualadaptation, nikolaidis_HRI_sharedautonomy, nikolaidis_IJRR_mutualadaptation}. Table~\ref{table:summary} organizes the works presented in this paper, with respect to the different aspects of the general formulation that each one addresses.

Each section articulates the different assumptions and explains how these lead to the robot behaviors that we observed in real-time interactions with  actual human subjects, in a variety of manufacturing, home environments and assistive care settings (Fig.~\ref{fig:applications}).
\\

\section{Problem Formulation} \label{sec:formulation}
Human-robot collaboration can be formulated as a \emph{two player game with partial information}. We let $x_t$ be the world state that captures the information that human and robot use at time $t$ to take actions $a^{\textrm{R}}_t$, $a^{\textrm{H}}_t$ in a collaborative task. Over the course of a task of total time duration $T$, robot and human receive an accumulated reward:

\begin{equation*}
\sum_{t=1}^T R^{\textrm{R}}(x_t,a^{\textrm{R}}_t, a^{\textrm{H}}_t)
\end{equation*}
 for the robot and  \\ 
\begin{equation*}
 \sum_{t=1}^T R^{\textrm{H}}(x_t,a^{\textrm{R}}_t, a^{\textrm{H}}_t)
\end{equation*}
 for the human. 

We assume a robot policy $\pi^{\textrm{R}}$, which maps world states to actions. The human chooses their own actions based on a human policy $\pi^{\textrm{H}}$. If the robot could control both its own and the human actions, it would simply compute the policies that maximize its own reward.

However, the human is not another actuator that the robot can control. Instead, the robot can only \em estimate \em the human decision making process from observation and \em make predictions \em about future human behavior, which in turn will affect the reward that the robot will receive. 

 Therefore, the optimal policy for the robot is computed by taking the expectation over human policies $\pi^{\textrm{H}}$.

\begin{align}
\pi^{{\mathrm{R}}^*} &\in \argmax{\pi^{\mathrm{R}}}\ { \mathbb{E}} \left[\sum_{t=1}^T  R^{\textrm{R}}(x_t, a^\mathrm{R}_t, a^\mathrm{H}_t)| \pi^{\mathrm{R}},\pi^{\mathrm{H}} \right]
\label{eq:main}
\end{align}

Solving this optimization is challenging: First, the human reward $R^{\textrm{H}}$ may be unknown to the robot in advance. Second, even if the robot knows $R^{\textrm{H}}$, it may be unable to predict accurately the human actions, since human behavior is characterized by bounded rationality~\cite{kahneman_American_bounded}. Third, even if the human acts always rationally, exhaustively searching for the equilibria is computationally intractable in most cases~\cite{papadimitriou_AGT_complexity}. Finally, even if $R^{\textrm{H}}\equiv R^{\textrm{R}}$, most real-world problems have multiple equilibria, and in the absence of a signaling mechanism, it is impossible to know which ones the agents will choose.

 \begin{figure*}[t!]
 \centering
 \includegraphics[width=1.0\textwidth]{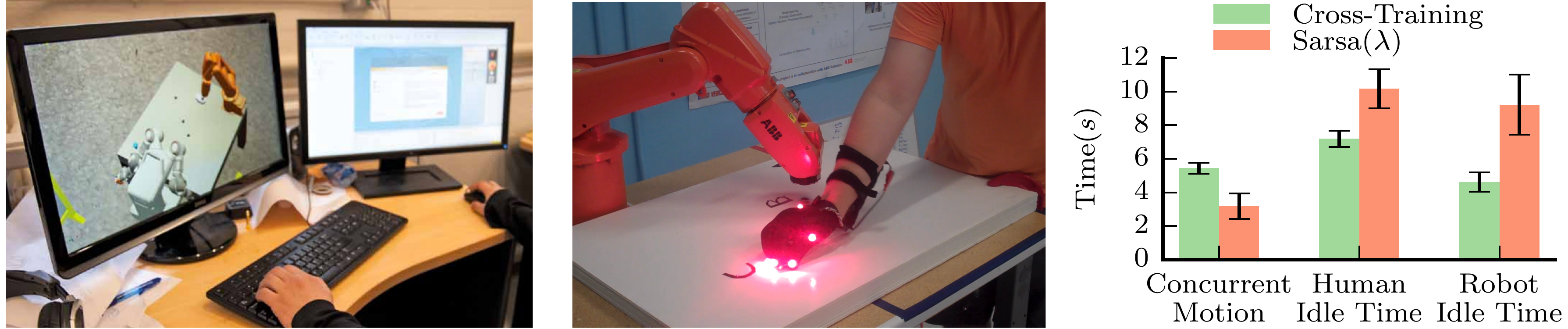}
 \caption{Cross-training in a virtual environment leads to fluent human-robot teaming.}
 \label{fig:cross-training}
 \end{figure*}

\section{Robot Adaptation.} \label{sec:robot_adaptation}
In several manufacturing applications, such as assembly tasks, although important concepts such as tolerances and completion times are well-defined, many of the details are largely left up to the individualized preference of the mechanics. A robotic assistant interacting with a new human worker should be able to learn the preferences of its human partner in order to be an effective teammate.  We assume a \emph{leader-assistant} teamwork model, where the human leader's preference is captured by the human reward function  $R^{\textrm{H}}$ and the human policy  $\pi^{\textrm{H}}$.  In this model, the goal of the robot is to execute actions aligned with the human preference. Therefore, in Eq.~\ref{eq:main}, we have:
\begin{equation*} 
R^{\textrm{R}} \equiv R^{\textrm{H}}
\end{equation*}

\noindent\textbf{Learning of a Human Model.}~Learning jointly $\pi^{\textrm{H}}$ and  $R^{\textrm{H}}$ can be challenging in settings where human and robot take actions simultaneously, and do not have identical action sets. To enable a robot to learn the human preference in collaborative settings, we looked at how humans communicate effectively their preferences in human teams. In previous work~\cite{shah2011improved}, insights from human teamwork have informed the design of a robot plan execution system which improved human-robot team performance. We focused on a team training technique known as \emph{cross-training}, where human team-members switch roles to develop shared expectations on the task. This, in turn allows them to anticipate one another's needs and coordinate effectively. Using this insight, we proposed human-robot cross-training~\cite{nikolaidis_IJRR_crosstraining, nikolaidis_HRI_crosstraining, nikolaidis_ISR_collaboration}, a framework where the robot learns a model of its human counter-part through two phases: a forward-phase, where human and robot follow their pre-defined roles, and a rotation phase, where the roles of human and robot are switched. The forward phase enables the robot to observe the human actions and estimate the human policy $\pi^{\textrm{H}}$. The rotation phase allows the robot to observe the human inputs on the robot actions and infer the human preference  $R^{\textrm{H}}$ from human demonstrations~\cite{argall2009survey}. After each training round, which occurs in a virtual environment, the robot uses the new estimates to compute the optimal policy from Eq.~\ref{eq:main}. Our studies showed that cross-training provides significant improvements in quantitative metrics of team fluency, as well as in the perceived robot performance and trust in the robot (Fig.~\ref{fig:cross-training}). These results provide the first indication that effective and fluent human-robot teaming may be best achieved by modeling effective training practices for human teamwork.

\begin{figure}[h!]
\centering
\includegraphics[width=1.0\linewidth]{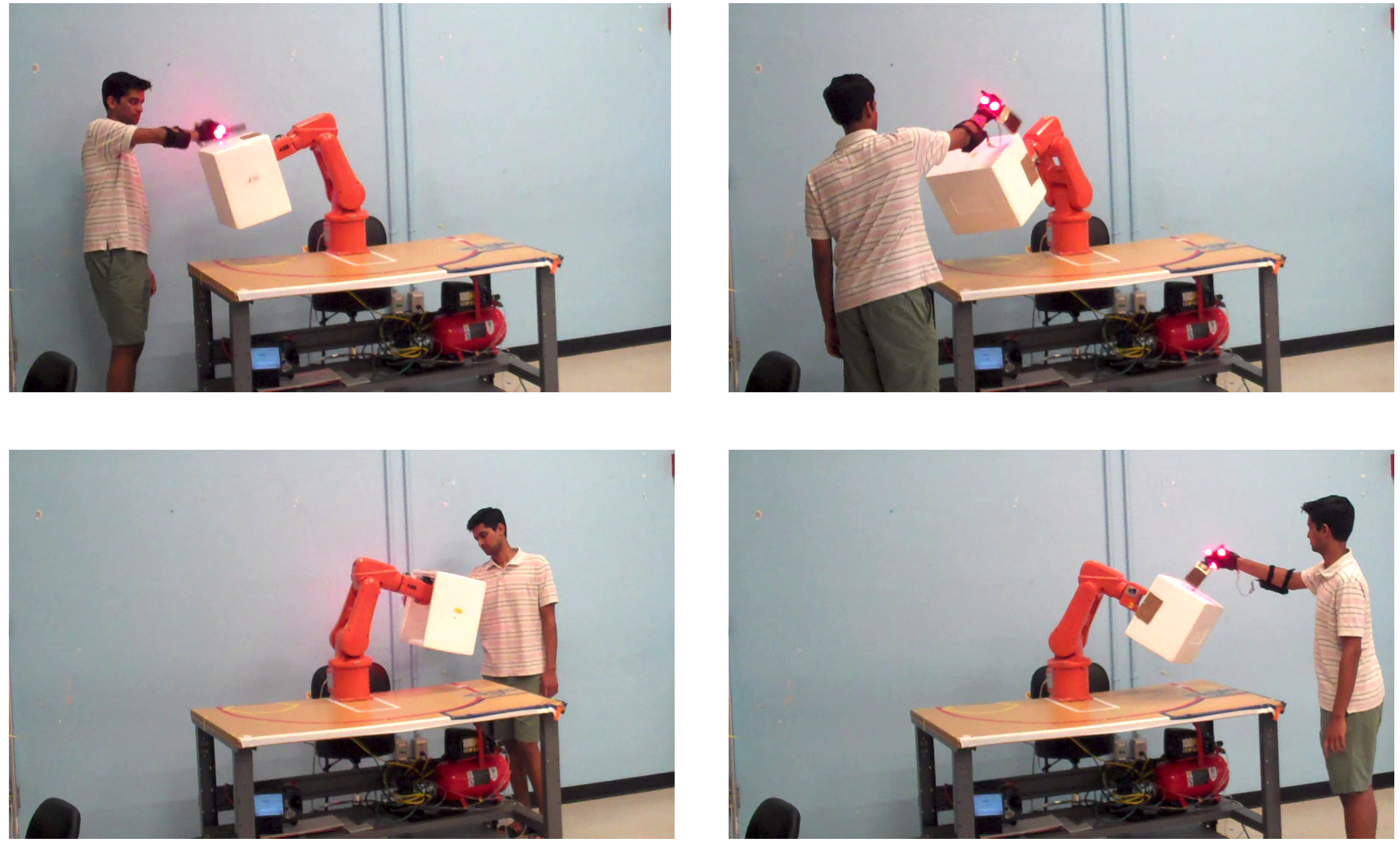} 
\caption{We clustered participants into types, based on how their preference of executing a hand-finishing task with the robot.}
\label{fig:human-types}
\end{figure}

\noindent\textbf{Inference of a Human Type.}~Cross-training works by learning an individualized model for each human teammate. For more complex tasks, this results in a large number of training rounds, which can be tedious from a human-robot interaction perspective. However, our pilot studies showed that even when there was a very large number of feasible action sequences towards task completion, people followed a limited number of ``dominant'' strategies. Using this insight, we used unsupervised learning techniques to identify distinct \emph{human types} from joint-action demonstrations (Fig.~\ref{fig:human-types})~\cite{nikolaidis_HRI_learningtypes}. For each type $y \in Y$, we used supervised learning techniques to learn the human reward  $R^{\textrm{H}}(x_t,a^{\textrm{R}}_t, a^{\textrm{H}}_t; y)$, as well as the human policy $\pi^{\textrm{H}}(x_t; y)$, This simplified the problem of learning  $R^{\textrm{H}}, \pi^{\textrm{H}}$  of a new human worker, to simply inferring their type $y$.  We enabled this inference by denoting the human type as a latent variable in a partially observable stochastic process (POMDP).  This allowed the robot to take information seeking actions in order to infer online the type a new user, and execute actions aligned with the preference of that type. This draws upon insights from previous work on cooperative games~\cite{macindoe2012pomcop} and vehicle navigation~\cite{bandyopadhyay2013intention}, where the human intent was modeled as a latent variable in a POMDP, albeit with prespecified models of human types. In a human subject experiment, participants found that the robot executing the computed policy anticipated their actions, and in complex robot configurations they completed the task faster than manually annotating robot actions.

\section{Human Adaptation.}  \label{sec:human_adaptation}
As robotics systems become more advanced, they have access to information that the human may not have; this suggests that rather than always following the human, they could use this information as \emph{equal partners} to guide their human teammates towards better ways of doing the task. 
In that case, it is no longer optimal for the robot to optimize the human reward function; instead, the robot should maximize its own reward function, which is different than the human's:
\begin{equation*} 
R^{\textrm{R}} \neq R^{\textrm{H}}
\end{equation*}
An improvement upon the leader-assistant setting is to recognize that the human policy can change based on the robot actions.  We let a history of world states and robot actions $h_t$:
\begin{equation*} 
 h_t = (x_{0}, a^{\textrm{R}}_{0}, ...,x_{t}, a^{\textrm{R}}_{t})
\end{equation*} 

Given this history, the human policy  $\pi^{\textrm{H}}(x_t, h_t; y)$ is a function not only of the current world state $x_t$ and human type $y_t$, but also of the history $h_t$. Modeling the human policy as a function of the robot actions and solving the optimization of Eq.~\ref{eq:main} makes the interaction an \emph{underactuated dynamical system}, where the robot reasons over how its own actions affect future human actions, and takes that into account into its own decision making.

\noindent\textbf{A Bounded-Memory Model.}~This history $h_t$ can grow arbitrarily large, making optimizing for the robot actions computationally intractable. In practice, however, people do not have perfect recall. Using insights from work on bounded-rationality in behavioral economics, we simplify the optimization, using a Bounded-memory human Adaptation Model (BAM)~\cite{nikolaidis_HRI_mutualadaptation, nikolaidis_IJRR_mutualadaptation}. The human observes a recent history of interactions $h_k$:
\begin{equation*} 
 h_k = (x_{t-k}, a^{\textrm{R}}_{t-k}, ...,x_{t}, a^{\textrm{R}}_{t})
\end{equation*} 
and infers a probability distribution over joint global plans that the robot is executing. They may then switch from the plan they currently follow, to a new plan demonstrated by the robot with some probability associated with their type $y$ -- their willingness to adapt to the robot.

\begin{figure}[t!]
\centering
\includegraphics[width=0.8\linewidth]{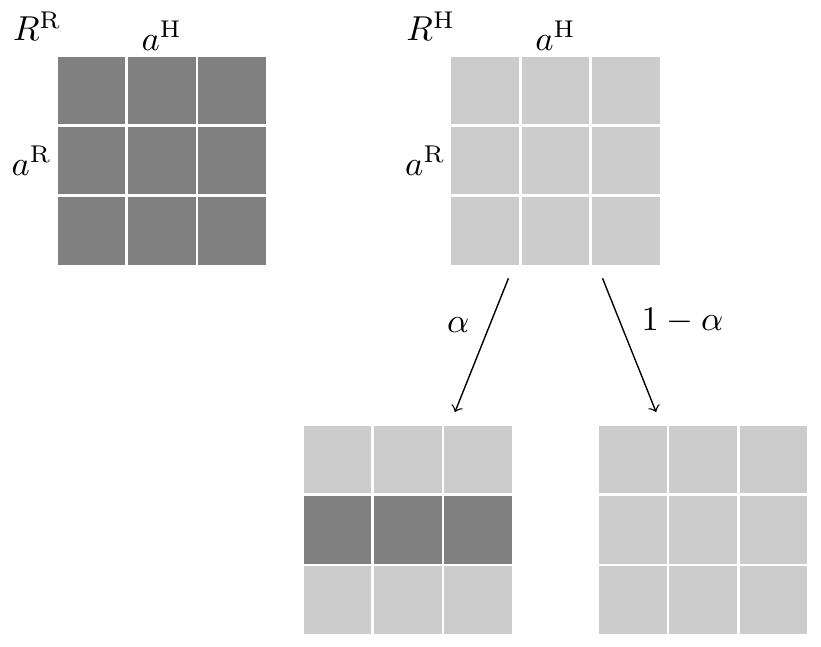} 
\caption{The human state changes, as the human plays with probability $\alpha$ the best response to the robot action $a^{\textrm{R}}$ (middle row) with respect to the robot reward function $R^{\textrm{R}}$, instead of playing the best-response with respect to their own reward function $R^{\textrm{H}}$.}
\label{fig:game-theory}
\end{figure}

\begin{figure*}[t!]
\centering
\setlength\tabcolsep{-1.0pt}
\centering
\begin{tabular}{ccc}
\begin{subfigure}[l]{0.33\linewidth}
\centering
\includegraphics[height=3.2cm]{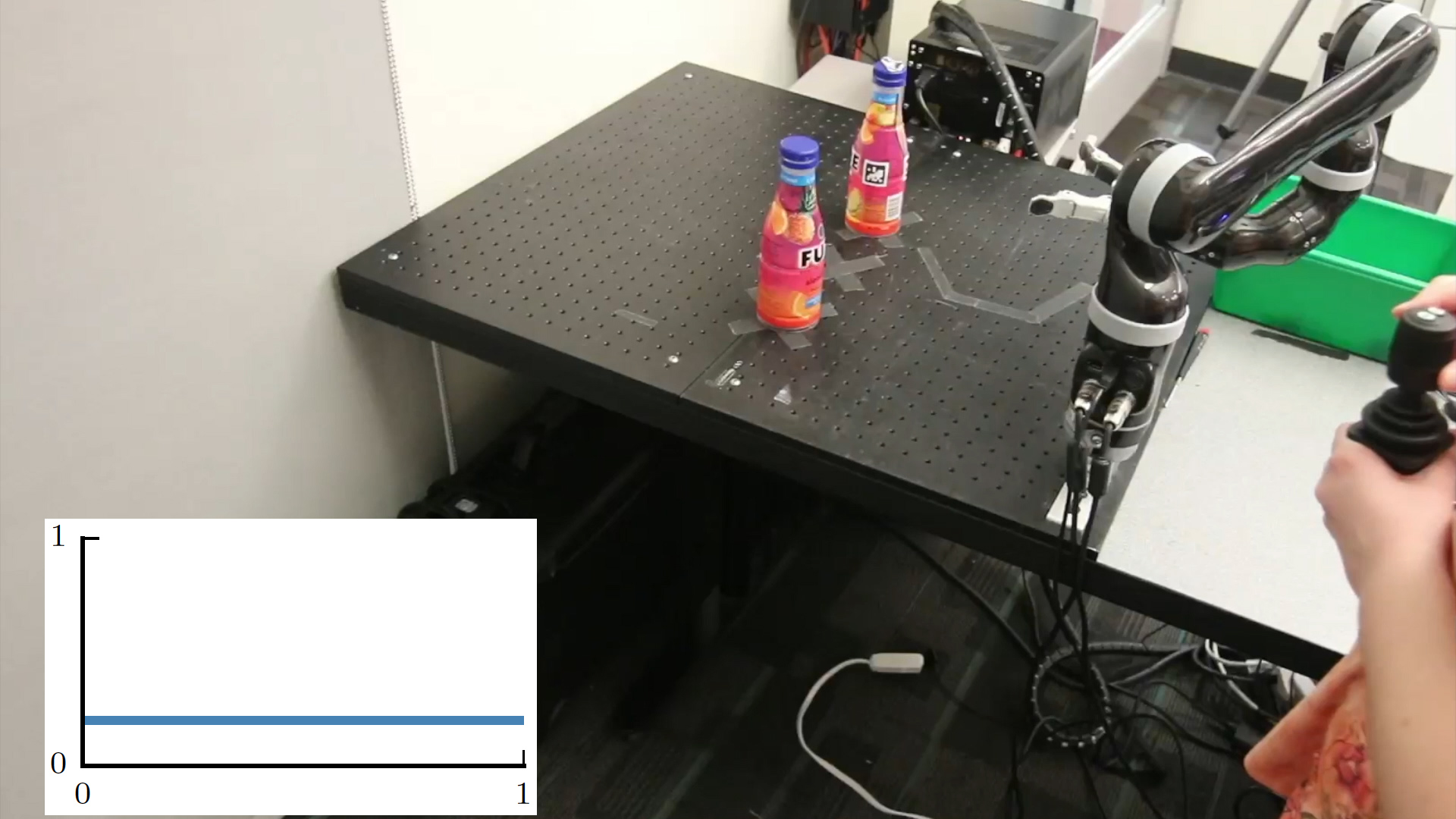} \\
\label{fig:path-left}
\end{subfigure}
&
\begin{subfigure}[l]{0.33\linewidth}
\centering
\includegraphics[height=3.2cm]{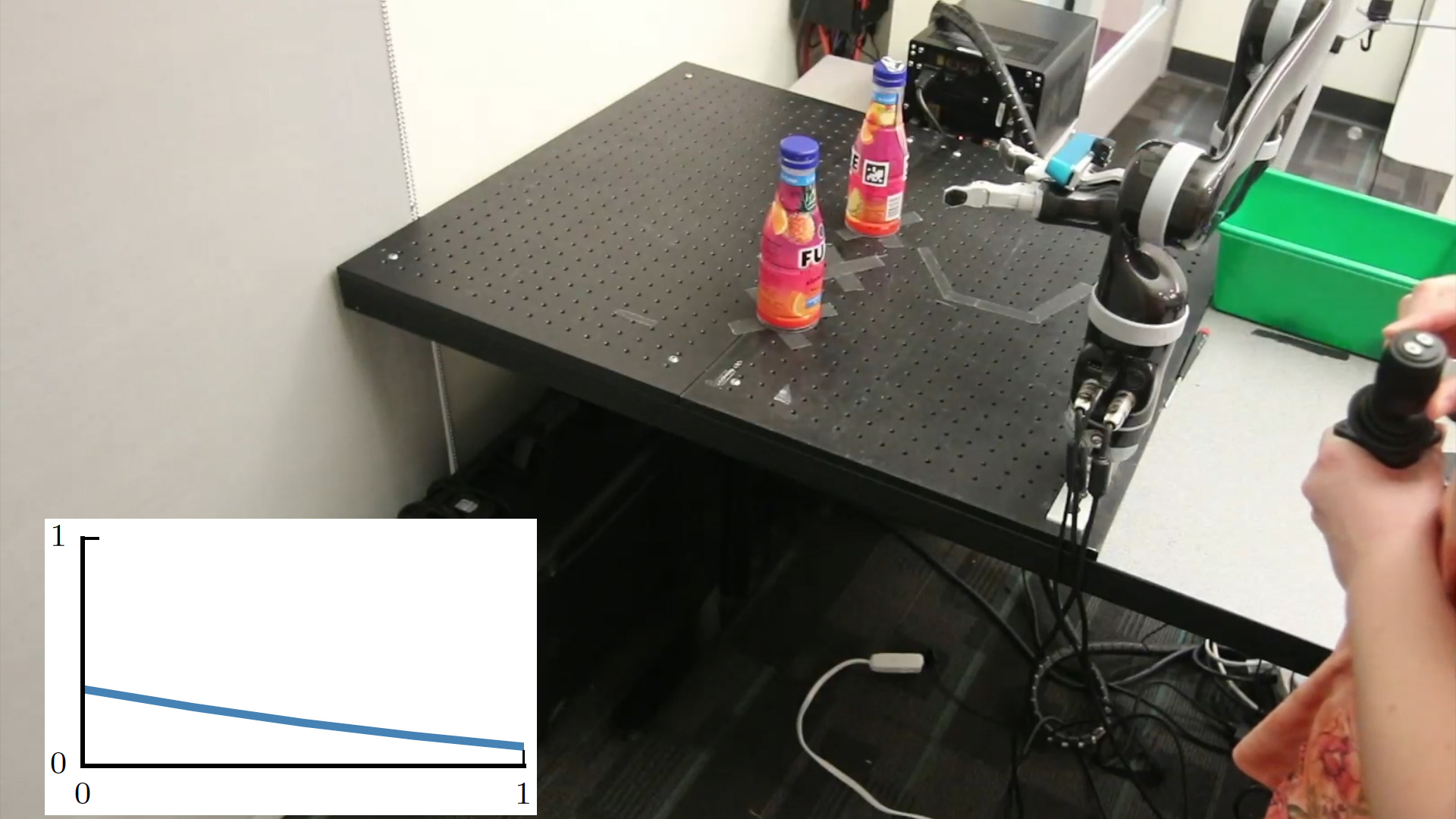} \\
\label{fig:path-left}
\end{subfigure}
&
\begin{subfigure}[l]{0.33\linewidth}
\centering
\includegraphics[height=3.2cm]{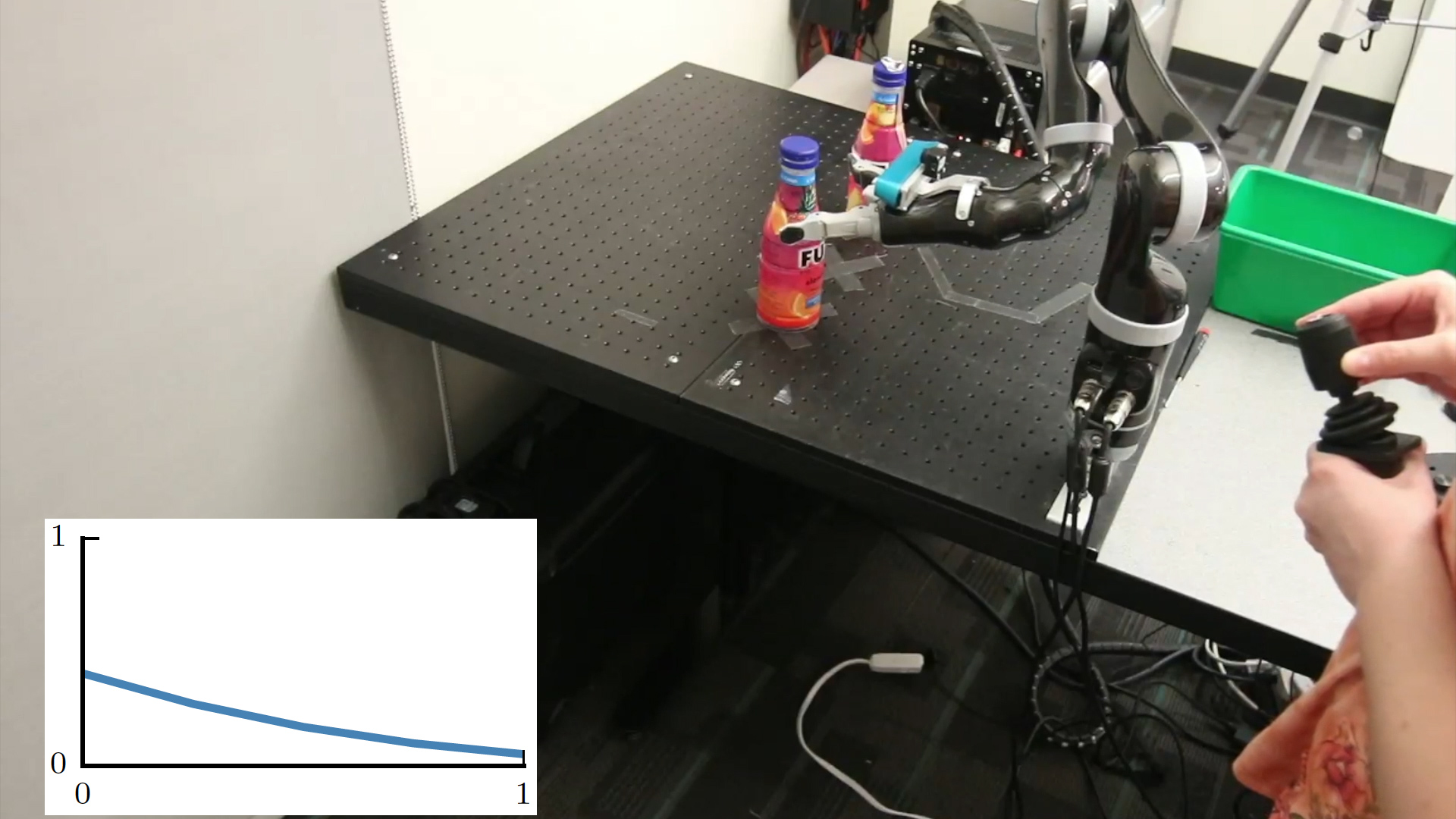} \\
\label{fig:path-right}
\end{subfigure}
\vspace{0.2cm}
\\
\begin{subfigure}[l]{0.33\linewidth}
\centering
\includegraphics[height=3.2cm]{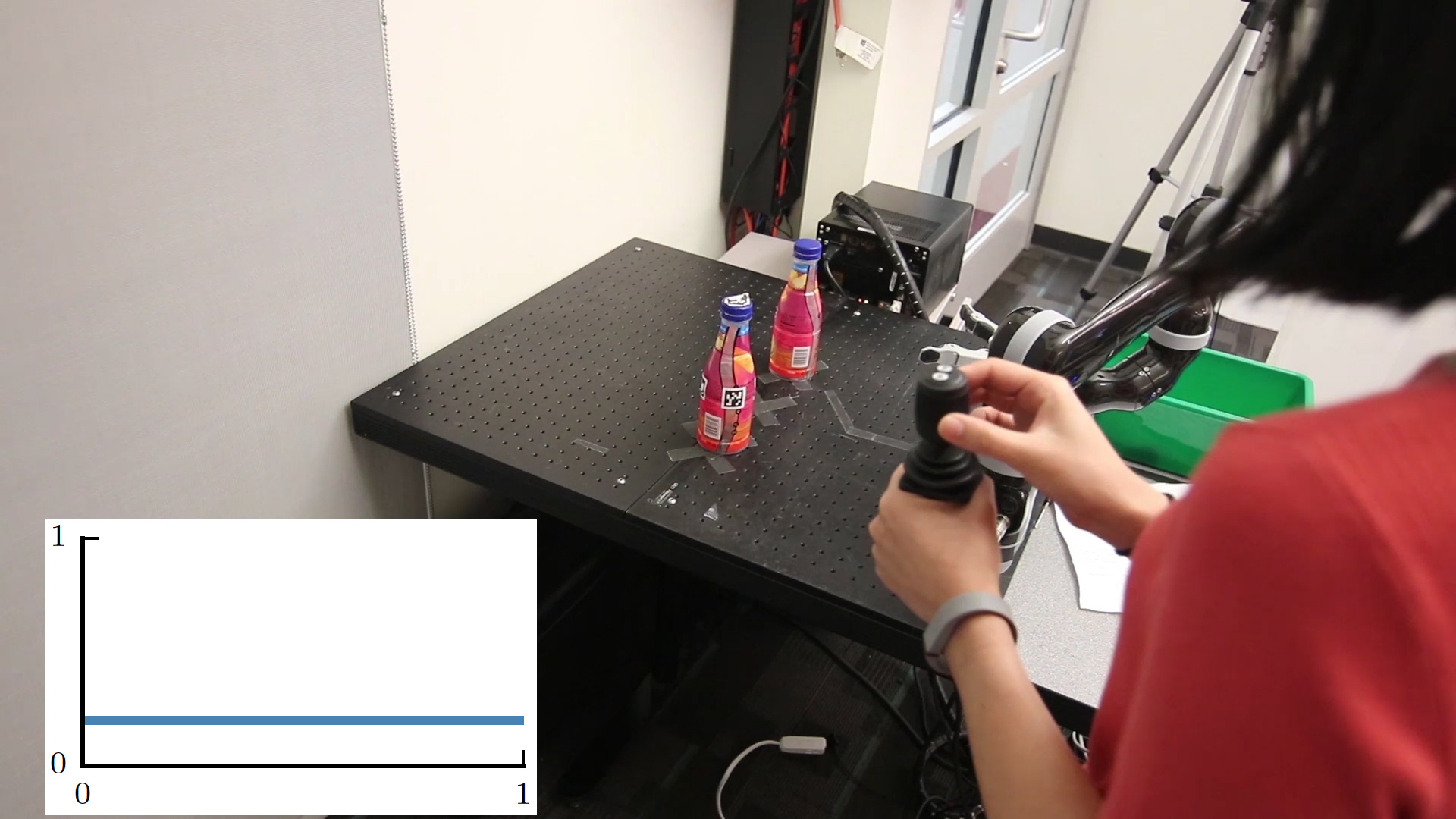} \\
\label{fig:straight-path}
\end{subfigure}
&
\begin{subfigure}[l]{0.33\linewidth}
\centering
\includegraphics[height=3.2cm]{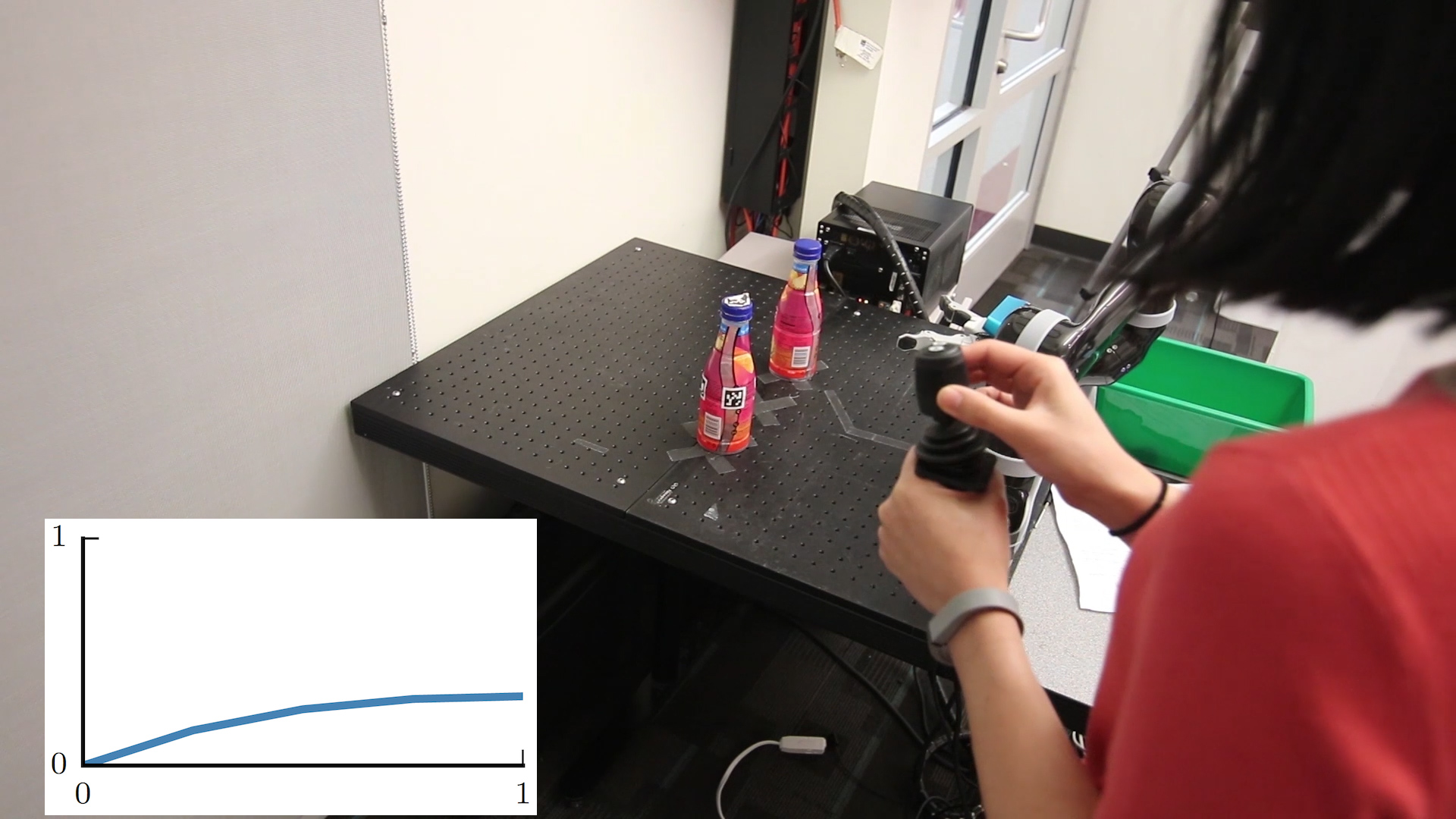} \\
\label{fig:straight-path}
\end{subfigure}
&
\begin{subfigure}[l]{0.33\linewidth}
\centering
\hspace{-0.24cm}
\includegraphics[height=3.2cm]{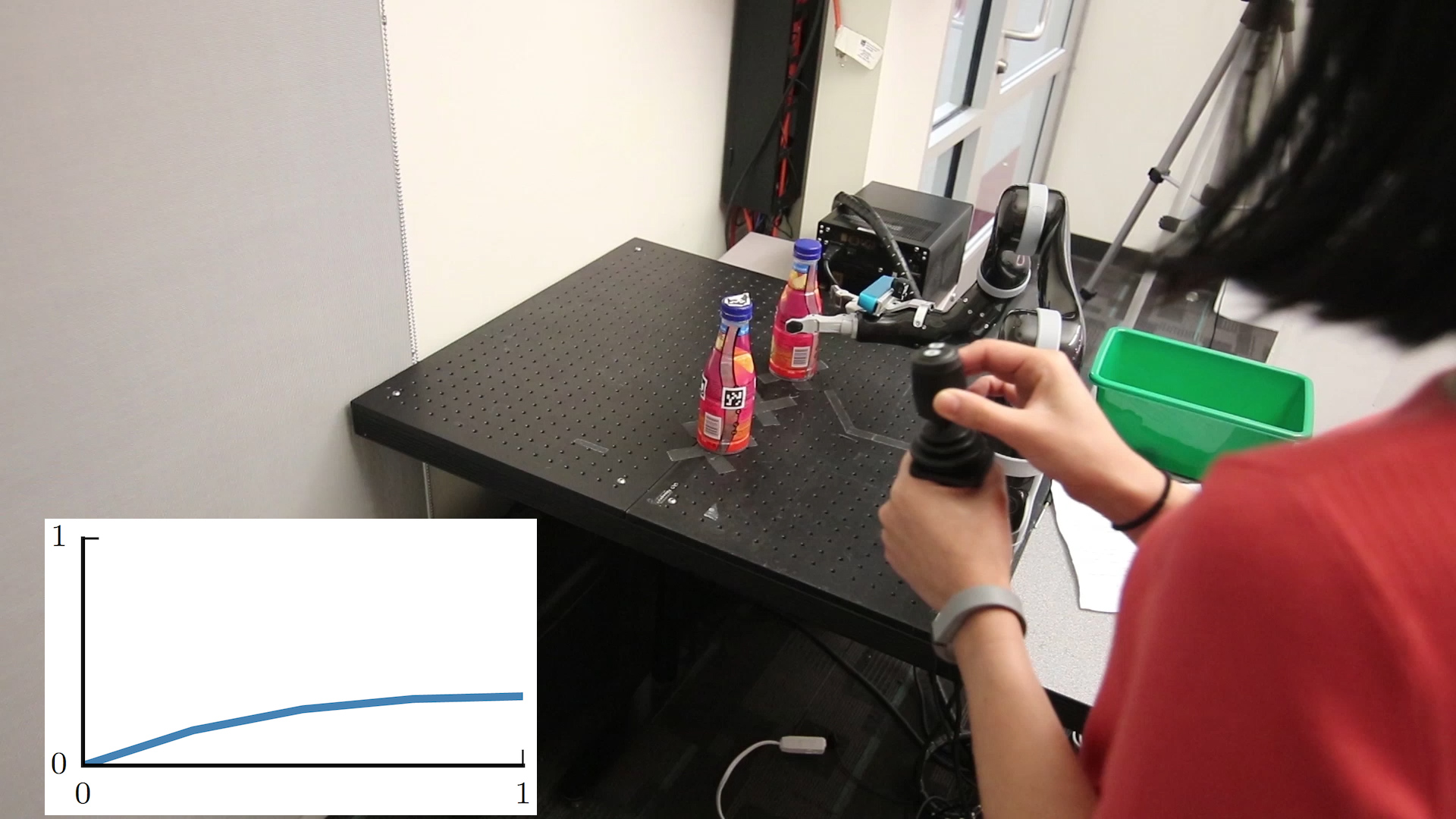} \\
\label{fig:straight-path}
\end{subfigure}
\end{tabular}
\caption{Two different users control an assistive arm through joystick inputs. The plot indicates the robot belief (y-axis) on user's adaptability (x-axis). The robot starts by moving straight towards both bottles, in order to infer the user adaptability. Top, left to right: A non-adaptable user insists on moving the joystick to the left and the robot adapts to the user to retain their trust. Bottom, left to right: The user changes their inputs to align them with the robot actions. The robot infers that the user is adaptable and guides them towards the optimal goal (right bottle).}
\label{fig:mutual-adaptation}
\end{figure*}

\noindent\textbf{A Best-Response Model.}~A particular instance of treating interaction as an underactuated system is modeling people as computing a \emph{best-response}  to the last robot action using their reward function $R^{\textrm{H}}$:
\begin{equation*}
\pi^{{\mathrm{H}}}(x_t, a^\mathrm{R}_t;y) \in \argmax{a^{\mathrm{H}}}~R^{\textrm{H}}(x_t, a^\mathrm{R}_t, a^\mathrm{H}_t;y)
\end{equation*}

In~\cite{nikolaidis_HRI_gametheory}, we additionally broke the assumption that $R^{\textrm{H}}$ is static; instead, as the human observes the robot's actions and the outcomes, they may change their reward function $R^{\textrm{H}}$ to match those of the robot's. We captured this change by assuming that the reward parameters $y$ are not static but change stochastically based on the robot actions, according to a transition function: $P(y_{t+1} |y_{t},a_t^{\textrm{R}})$. For instance, the human may learn the true best response to a robot action with some probability (Fig.~\ref{fig:game-theory}). 
 Including this model in the optimization of Eq.~\ref{eq:main} allows the robot to take \emph{informative} actions to teach the human, even by failing in purpose~\cite{nikolaidis_HRI_gametheory}.

 The cardinality of $Y$ can grow exponential to the number of robot actions, which makes the problem computationally hard. Using our intuition that  early exploration should be encouraged, in~\cite{nikolaidis_HRI_gametheory} we simplified the computation via a \emph{linear-time optimal algorithm} for solving the maximization of Eq.~\ref{eq:main}, and showed that our model can capture accurately the evolution of human behavior in a table-clearing setting. 

\section{Mutual Adaptation.} \label{sec:mutual_adaptation}
In our models of human adaptation, we have assumed that the robot knows the type $y$ of the human, which parameterizes the human policy $\pi^{\textrm{H}}$ and the human reward function  $R^{\textrm{H}}$. However, our studies have shown that there is a large variability among different types $y$. Additionally, the type $y$ of a new human worker is typically unknown in advance to the robot and it cannot be fully observed. In~\cite{nikolaidis_HRI_mutualadaptation, nikolaidis_HRI_sharedautonomy, nikolaidis_IJRR_mutualadaptation} we relax the assumption of a known $y$ for the human. Instead, we treat $y$ as a latent variable in a partially observable stochastic process, in particularly a mixed-observability Markov decision process, which has been shown to achieve significant computational efficiency~\cite{ong_IJRR_planning}.  This allows the robot to take information seeking actions to infer online the parameter $y$, which specifies how the human policy $\pi^{\textrm{H}}$ is affected by the robot's own actions. As a result, human and robot \emph{mutually adapt} to each other; the robot builds online a model of how the human adapts to the robot by inferring their type $y$, and adapts its own actions in return.

\begin{figure}[t!]
\centering
\includegraphics[width=0.6\linewidth]{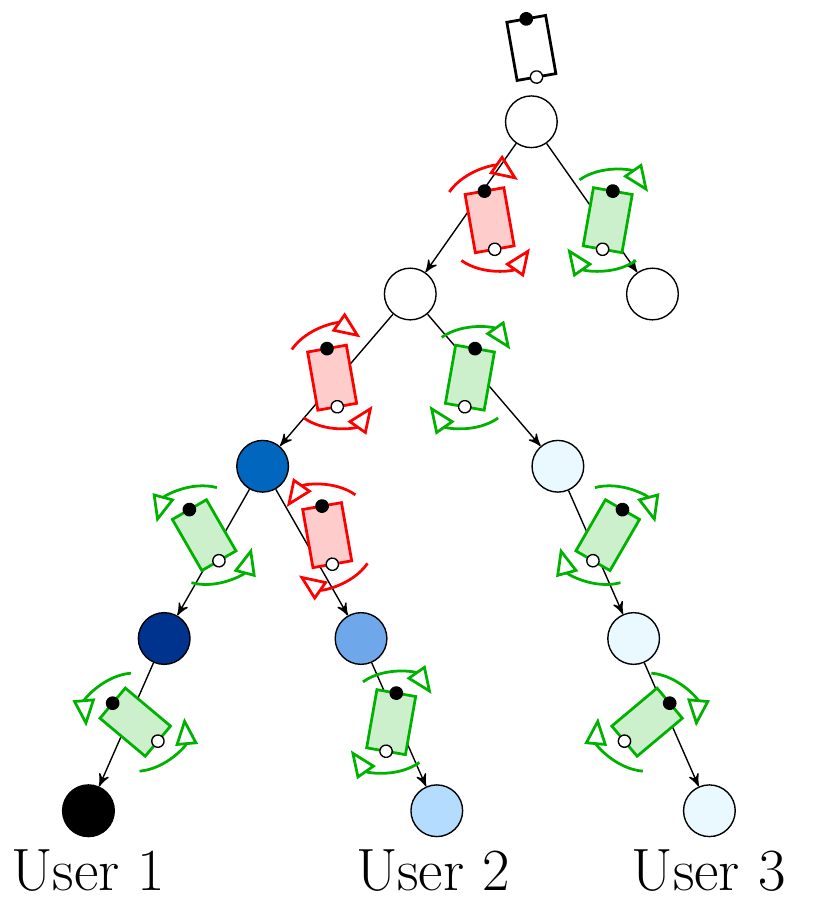} 
\caption{Different paths on MOMDP policy tree for a human-robot (white/black
dot) table-carrying task. The circle color represents the belief on the human adaptability, with darker shades indicating higher probability for smaller values (less adaptability). The white circles denote a uniform distribution over human adaptability. The table color indicates disagreement (red) or agreement (green) between human and robot actions. User 1 is inferred as non-adaptable, whereas Users 2 and 3 are adaptable.}
\label{fig:tree}
\end{figure}

\begin{figure}[t!]
\centering
\includegraphics[width=1.0\linewidth]{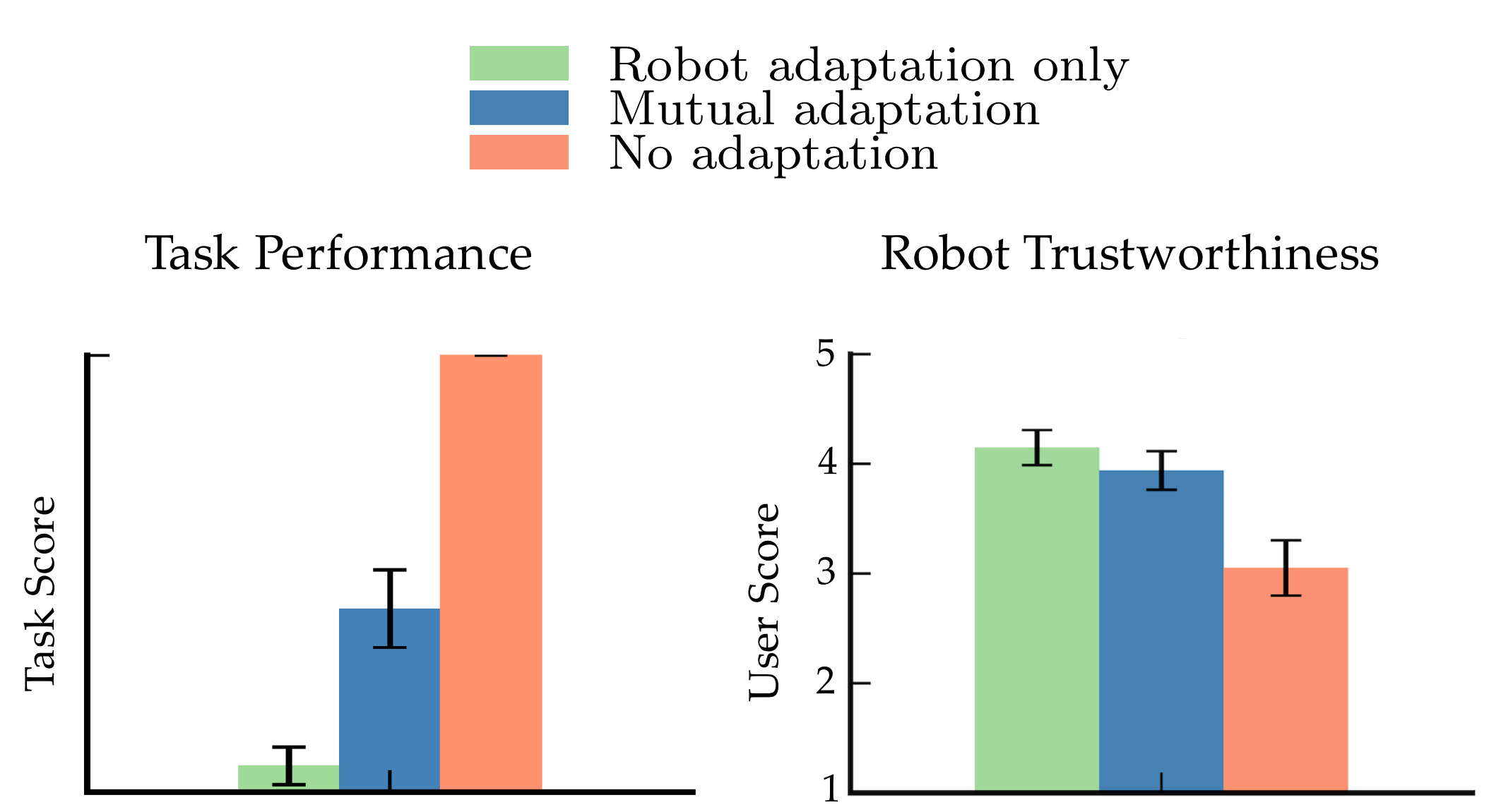} 
\caption{Results for the three conditions in our human subjects experiment~\cite{nikolaidis_HRI_sharedautonomy}. Robot-adaptation only: the robot assisted the user on their preference. Mutual-adaptation: The proposed formalism of Sec.~\ref{sec:mutual_adaptation}. No adaptation: The robot goes to the optimal goal always ignoring the user. Mutual-adaptation performs better than robot adaptation only, since a large number of users adapted to the robot, while achieving comparable trust-ratings. No adaptation achieves the maximum performance by definition, but to the detriment of reported user trust in the robot.}
\label{fig:shared-autonomy-results}
\end{figure}

We instantiated the mutual adaptation formalism using as  human type $y$ the human adaptability, that is their willingness to adapt to new strategy demonstrated the robot. Fig.~\ref{fig:mutual-adaptation} shows an instance of the mutual adaptation formalism in a shared autonomy setting. In this setting, the robot infers a distribution over goals based on the user input, and assists the user for that distribution~\cite{javdani2017shared}. Mutual adaptation additionally allows the robot to reason over how the user goal may change, as the user adapts themselves to the robot. In our example, the starting user preference is for the robot to go towards the left bottle, while the robot considers the right bottle as a better goal, for instance because it is heavier and it should be placed in the green bin first. The robot starts by moving straight (information-seeking action), observing whether the human changes their inputs to follow the robot (adaptable user), or insists on pushing the joystick to the left (non-adaptable). If the human is adaptable (bottom-row), the robot will guide them towards a better way of doing the task. If the human is non-adaptable (top-row), the robot follows the human preference towards the left-bottle, in order to retain their trust. Similarly, Fig.~\ref{fig:tree} shows the MOMDP policy tree for a table-carrying task. The robot infers the user adaptability, and chooses whether to move the table following the user's initial preference, or to guide the user towards moving the table in the opposite direction. Our experiments in collaborative manipulation~\cite{nikolaidis_HRI_mutualadaptation}, social navigation~\cite{nikolaidis_IJRR_mutualadaptation} and shared autonomy~\cite{nikolaidis_HRI_sharedautonomy} settings showed that, when compared to one-way robot adaptation to the human, mutual adaptation significantly improved human-robot team performance, while achieving comparable trust ratings (Fig.~\ref{fig:shared-autonomy-results}).

\section{Discussion}
We presented a summary of our work in human-robot collaboration. We formulated the general problem as a two-player game with incomplete information, where human and robot know each other's goals. We then made a set of different assumptions and approximations within the scope of this general formulation. Each assumption resulted in diverse and exciting team coordination behaviors, which had a strong effect on team performance.

We have shown that representing the human preference as a human reward function unknown to the robot and computing the robot policy that maximizes this function results in robot adaptation to the human. Assuming the human reward function to be known and treating the interaction as an underactuated dynamical system results in human adaptation to the robot. Closing the loop between the two results in mutual adaptation, where the robot builds online a model of human adaptation, and adapts its own actions in return. 

 We have applied the mutual adaptation formalism in collaborative manipulation, social navigation and shared autonomy settings. We have also been using the same formalism to incorporate verbal communication between the robot and the human~\cite{nikolaidis_JHRI_verbal}. We are excited about generalizing our work in a variety of domains, robot morphologies and interaction modalities, where an autonomous system plans its actions by incorporating the human internal state. The number of applications is vast: an autonomous car can infer the aggressiveness of a nearby driver and choose to wait or proceed; a GPS system may infer whether a user is willing to follow its prompts; a personal robot at home can ``nudge'' a user about taking breaks and sleeping more. 

 As these applications become more complex, our work has a number of limitations. The models of human internal state that robots can build reliably are restricted and achieving optimal behavior in large, high-dimensional spaces faces computational intractability. To this end, flexible, compact representations of the human internal state and new algorithms for reasoning about these representations give much promise. 

Overall, we believe that we have brought about a better understanding of different ways that probabilistic planning and game-theoretic algorithms can support principled reasoning in robotic systems that collaborate with people. We look forward to continue addressing the exciting scientific challenges in this area. 

 \bibliographystyle{unsrt}
 \bibliography{conferences,journals}

\end{document}